\documentclass[runningheads]{llncs}

\usepackage{eccv}

\usepackage{eccvabbrv}

\usepackage{graphicx}
\usepackage{booktabs}

\usepackage{algorithm}
\usepackage[noend]{algpseudocode}
\usepackage{amsmath}

\usepackage{multirow}

\usepackage[accsupp]{axessibility}  % Improves PDF readability for those with disabilities.

\usepackage{hyperref}

\usepackage{orcidlink}

\begin{document}

\title{Dynamic Retraining-Updating Mean Teacher for Source-Free Object Detection} 

\titlerunning{Dynamic Retraining-Updating Mean Teacher for SFOD}

\author{Trinh Le Ba Khanh\orcidlink{0000-0003-1688-9003} \and
Huy-Hung Nguyen\orcidlink{0000-0001-5394-9381} \and
Long Hoang Pham\orcidlink{0000-0002-3240-657X} \and \\
Duong Nguyen-Ngoc Tran\orcidlink{0000-0001-7537-6377} \and
Jae Wook Jeon\orcidlink{0000-0003-0037-112X}\thanks{~Corresponding author}}

\authorrunning{T.~L.~B.~Khanh et al.}

\institute{Department of Electrical and Computer Engineering, Sungkyunkwan University\\
\email{\{trinhlbk,huyhung91,phlong,duongtran,jwjeon\}@skku.edu}}

\maketitle

\begin{abstract}
In object detection, unsupervised domain adaptation (UDA) aims to transfer knowledge from a labeled source domain to an unlabeled target domain. However, UDA's reliance on labeled source data restricts its adaptability in privacy-related scenarios. This study focuses on source-free object detection (SFOD), which adapts a source-trained detector to an unlabeled target domain without using labeled source data. Recent advancements in self-training, particularly with the Mean Teacher (MT) framework, show promise for SFOD deployment. However, the absence of source supervision significantly compromises the stability of these approaches. We identify two primary issues, \emph{(1)} uncontrollable degradation of the teacher model due to inopportune updates from the student model, and \emph{(2)} the student model's tendency to replicate errors from incorrect pseudo labels, leading to it being trapped in a local optimum. Both factors contribute to a detrimental circular dependency, resulting in rapid performance degradation in recent self-training frameworks. To tackle these challenges, we propose the Dynamic Retraining-Updating (DRU) mechanism, which actively manages the student training and teacher updating processes to achieve co-evolutionary training. Additionally, we introduce Historical Student Loss to mitigate the influence of incorrect pseudo labels. Our method achieves state-of-the-art performance in the SFOD setting on multiple domain adaptation benchmarks, comparable to or even surpassing advanced UDA methods. The code will be released at \url{https://github.com/lbktrinh/DRU}.

% In this way, the dynamic teacher can integrate knowledge from past periods, effectively reducing error accumulation and enabling a more stable training process within the based framework. 
\end{abstract}

\section{Introduction}
\label{sec:introduction}
Unsupervised Domain Adaptation (UDA) has garnered significant attention in object detection \cite{Mttrans, Sfa, Dafaster, Swfaster, O2net}, focusing on transferring knowledge from source domains to target domains. UDA methods typically require access to both labeled source and unlabeled target data simultaneously to achieve feature alignment across domains. However, practical scenarios often involve privacy concerns, limiting access to source domain data and thereby introducing new challenges for UDA. In response, Source-Free Object Detection (SFOD) \cite{Sed, Pets, Lods, Irg, A2sfod} has recently emerged as a promising research direction. SFOD aims to adapt source-trained models to the unlabeled target domain without requiring access to labeled source data.

\begin{figure}[tb]
  \centering
  \includegraphics[width=1.0\columnwidth]{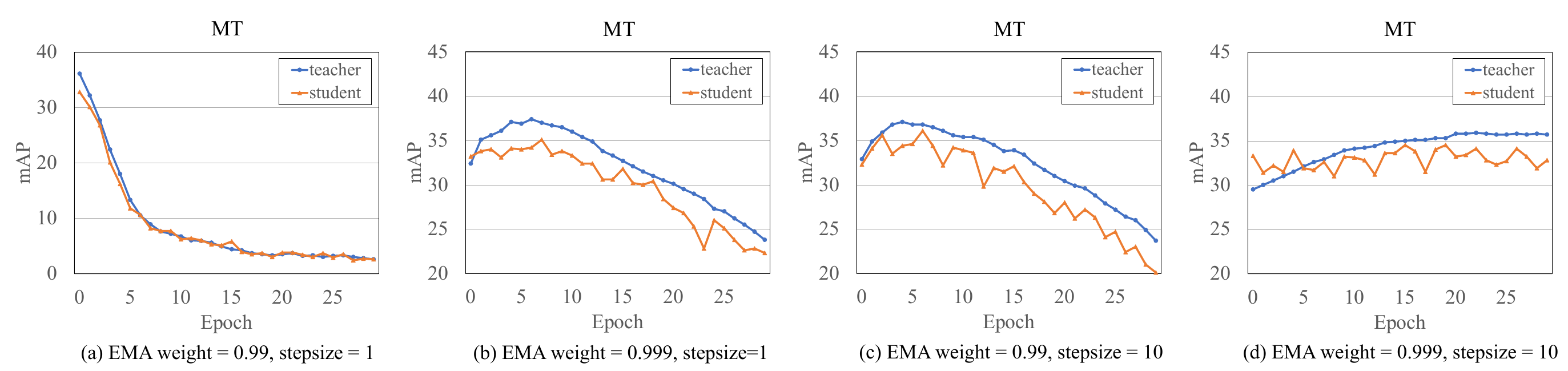}
  \caption{The training curves of different Mean Teacher training strategies on the validation set of Cityscapes $\to$ Foggy Cityscapes under the SFOD setting. These varied strategies consistently show a degradation phenomenon: the teacher model gradually degrades due to inappropriate updates from the student model, while the student model experiences performance deterioration due to inaccurate pseudo labels.
  }
  \label{fig:curve}
\end{figure}

In SFOD, the absence of source supervision makes self-training using pseudo labels conventional in this setting. Recent developments in self-training utilizing the Mean Teacher (MT) \cite{Mt} have shown great potential \cite{A2sfod, Irg, Lods, Hcl}. These MT-based methods employ a teacher model to guide the student model. The iterative process involves updating the teacher model via the Exponential Moving Average (EMA) weight of the student model, followed by training the student model using pseudo labels generated by the updating teacher model. Despite some progress, iterative self-training approaches still encounter decline issues due to the lack of labeled source data and noisy pseudo labels arising from domain shifts.

We investigate two factors contributing to the deterioration of self-training within the MT framework in SFOD: \emph{(1)} Inappropriate updates of the teacher model can result in susceptibility to accumulating errors, as it continuously aggregates improper knowledge from the unevolved student model; \emph{(2)} When the teacher model makes errors and provides incorrect information, the student model reproduces these mistakes without any correction, resulting in being stuck at a local optimum. Both factors establish a harmful cyclical relationship, where the teacher model offers poor guidance to the student model, and the student model provides biased knowledge to the teacher model. This cycle ultimately leads to the destruction of the self-training paradigm due to the absence of purposeful evolution. As illustrated in \cref{fig:curve}, adjusting the EMA weights or updating interval \cite{Mt, A2sfod, Irg} does not entirely solve the uncontrollable degradation in the self-training MT framework. Moreover, searching for an optimal EMA hyper-parameter for model updating is time-consuming.

\begin{figure}[tb]
  \centering
  \includegraphics[width=0.8\columnwidth]{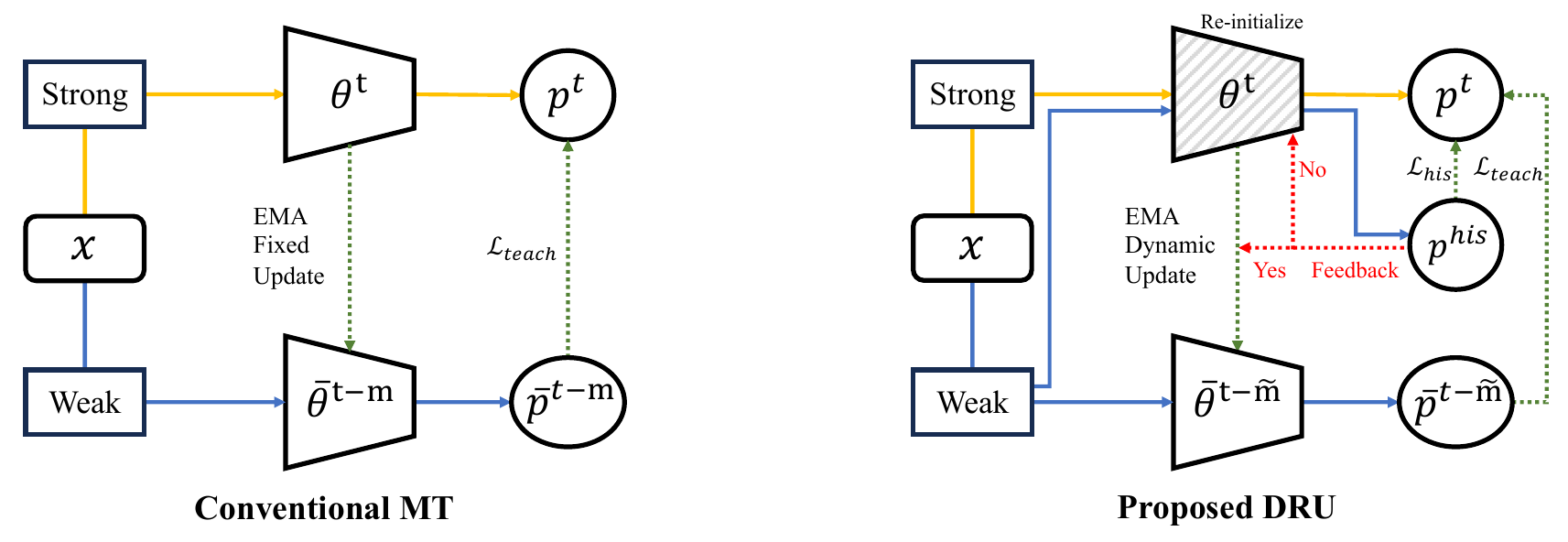}
  \caption{The comparison of the conventional Mean Teacher (MT) framework (\emph{left}) and our Dynamic Retraining-Updating (DRU) method (\emph{right}). {\bf Left:} In MT, the teacher model is continuously updated by a fixed interval $m$ ($m=1$ \cite{Mt} or $m=s, (s>1)$ \cite{A2sfod, Irg}). {\bf Right:} In DRU, the student model is dynamically retrained and the teacher model is dynamically updated based on prediction feedback. Additionally, the current student model is further supervised by the historical student model.
  }
  \label{fig:schema}
\end{figure}

In this study, we propose the Dynamic Retraining-Updating (DRU) method to address the aforementioned instability. The overview of DRU is depicted in \cref{fig:schema}. DRU actively manages student training and teacher updating processes to overcome the harmful interdependency inherent in the current self-training MT paradigm. Specifically, guided by the feedback on the student's evolutionary state, the student is dynamically retrained if trapped in a local optimum, while the teacher model is dynamically updated based on the evolved, retrained student. With this mechanism, the student can get out of sub-optimal states biased to inaccurate pseudo labels, and the teacher can accumulate valuable insights from the evolved student,  promoting co-evolution in training. Furthermore, we introduce an additional Historical Student Loss to mitigate rapid performance degradation caused by incorrect pseudo labels. This loss leverages knowledge from the historical student model to provide further supervision to the current student. With additional guidance from this loss, the student model can reduce the influence of incorrect pseudo labels generated by the deteriorating teacher model, resulting in more stable training.
 
Through efforts, DRU achieves state-of-the-art (SOTA) results across diverse SFOD scenarios, comparable to or even surpassing advanced UDA methods. Extensive experiments demonstrate that DRU effectively addresses the degradation problem inherent in self-training within the MT framework, leading to significant stability and adaptability during training. Our contributions can be summarized:
\begin{itemize} 
\item We explore the deterioration issue in the self-training MT-based framework, particularly when employed with transformer-based detectors, which are susceptible to domain shifts. To our knowledge, this is the first study to investigate the effectiveness of transformer-based detectors for SFOD problems.

\item We introduce the Dynamic Retraining-Updating (DRU) method to address the degradation observed in the MT framework. DRU actively controls student training and teacher updating to promote co-evolutionary training.

\item We propose the Historical Student Loss to prevent severe performance decline caused by noisy pseudo labels.

\item On multiple benchmarks, our DRU achieves SOTA performance compared to recent SFOD methods and surpasses several advanced UDA methods. 
\end{itemize}

\section{Related Works}
\label{sec:related-works}
\subsection{Object Detection} 
Object detection is a fundamental task in computer vision. Over the years, CNN-based detectors have attracted extensive attention \cite{Fasterrcnn, Fpn, Yolo, Fcos}. However, their reliance on hand-designed components, \eg non-maximum suppression, limits their ability to be trained end-to-end. Transformer-based detectors \cite{Detr, Defdetr, Dino} have recently emerged as an alternative for end-to-end object detection. While these detectors demonstrate robustness when applied to data from their training distribution, they experience degradation when faced with domain shifts. Our objective is to enhance the model's adaptability to new domains without requiring annotations. In this study, we adopt Deformable DETR \cite{Defdetr} as our baseline detector due to its streamlined simplicity and the adaptable transfer-learning capability of the attention mechanism.

% \subsection{Unsupervised Domain Adaptation}
% Unsupervised Domain Adaptation (UDA) for object detection tasks is typically categorized into three types. \emph{(1) Domain translation} aims to convert a target domain into a source-like domain using statistical information \cite{Alpher02, Alpher03} or translation models \cite{Afan, Umt, Dam}.
% \emph{(2) Adversarial learning} attempts to minimize the discrepancy between source and target domains within the feature space, either through domain discriminator \cite{Sfa} or adversarial loss functions \cite{Alpher02, Alpher03}. \emph{(3) Self-training} with pseudo labeling technique \cite{Pseudo} has been a dominant trend in UDA tasks. This approach is primarily built upon the mean-teacher (MT) framework \cite{Mt}, which utilizes pseudo-labels generated by the teacher model to guide the student model. However, reliance on labeled source data limits the utility of the MT approach in privacy-sensitive scenarios. In this study, our objective is to enhance the self-training paradigm using the MT framework in scenarios where source data is unavailable.

\subsection{Source-Free Object Detection}
Source-Free Object Detection (SFOD) is introduced to address adaptation scenarios involving privacy concerns. Without source domain supervision, applying existing UDA methods \cite{Dafaster, Sfa, Mttrans} to SFOD may lead to undesirable results, prone to significant instability during training. SED \cite{Sed} represents a pioneering attempt to introduce pseudo label self-training through self-entropy descent. A$^2$SFOD \cite{A2sfod} integrates an adversarial component into an MT framework to align feature spaces of source-similar and source-dissimilar images. IRG \cite{Irg} introduces a new contrastive loss guided by the Instance Relation Graph network to enhance target domain representations. To ensure stable self-training, HCL \cite{Hcl} constrains current models to prioritize knowledge aligned with historical models. PETS \cite{Pets} proposes a multi-teacher framework consisting of a static teacher, a dynamic teacher, and a student model to address self-training instability. While most SFOD methods leverage the MT framework \cite{Mt}, they overlook the essential co-evolution between student and teacher models, which is critical for stabilizing self-training paradigms. Based on observations, we introduce the Dynamic Retraining-Updating method to promote the improvement of both student and teacher models through training progress.

\subsection{Selective Retraining}
Due to the large number of parameters, optimizing a transformer-based model is prone to over-fitting on a small training set and getting stuck in a local optimum. In SFOD, this problem becomes worse due to incorrect pseudo labels. Recent approaches suggest that retraining provides an efficient solution to address this issue \cite{Siri, Mrt}. MRT \cite{Mrt} adopts periodic retraining within the MT framework to alleviate the impact of incorrect pseudo labels in UDA tasks. However, they rely on labeled source data to generate optimal initialization weights. We propose a new approach that dynamically retrains the student model to converge to a better local minimum when source data is unavailable.

\begin{figure}[tb]
  \centering
  \includegraphics[width=1.0\columnwidth]{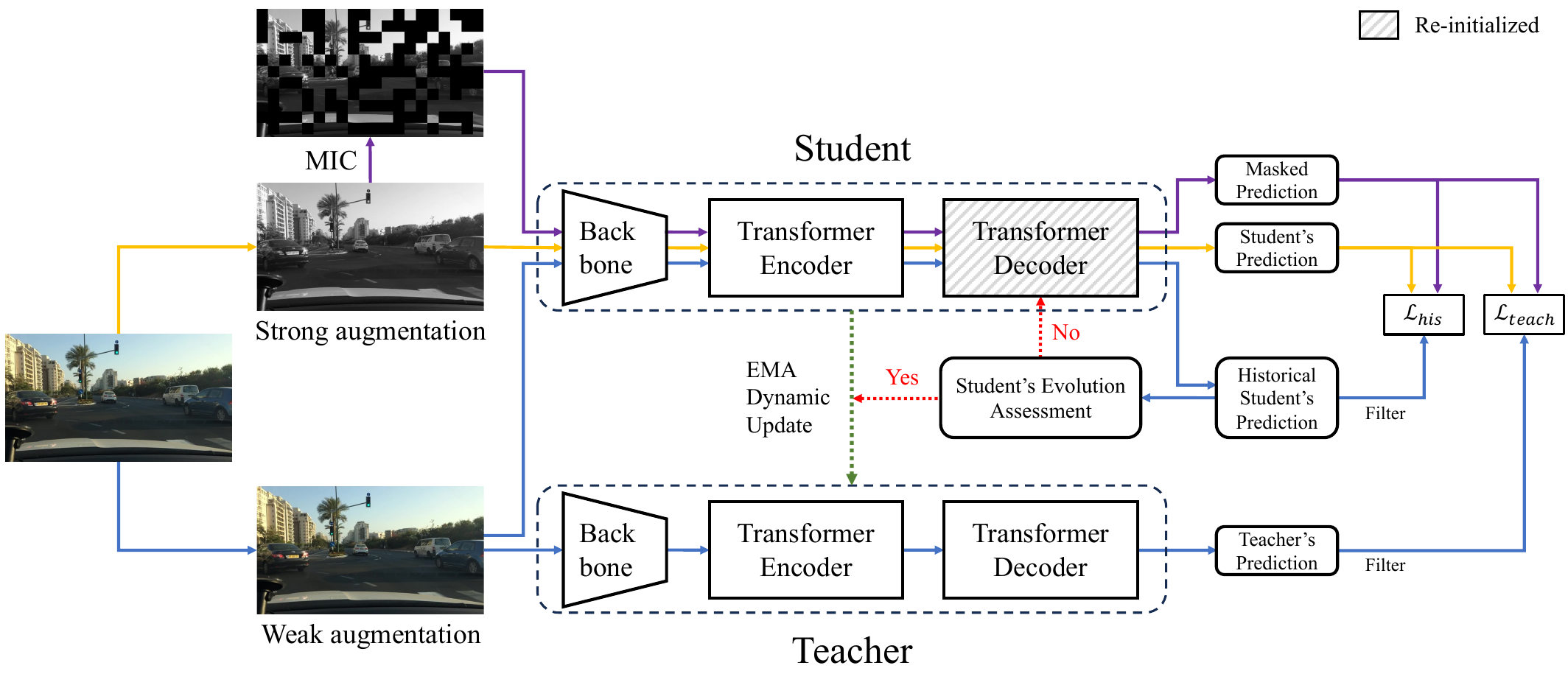}
  \caption{Overview of our Dynamic Retraining-Updating (DRU) method, which is built upon the Mean Teacher framework. DRU employs dynamic retraining of the student model and dynamic updating of the teacher model based on the student's evolution assessment. The student model is further supervised by Historical Student Loss $\mathcal {L}_{his}$. The details of the training process will be presented in \cref{alg:dru}}
  \label{fig:proposed}
\end{figure}

\section{Preliminary}
\label{sec:preliminary}
\subsection{Problem Definition} 
In this section, we define the problem formulation of SFOD. Let $D_s = \{(x_s, y_s)\}$  denote the labeled source domain data, and $D_t = \{x_t\}$ represent the unlabeled target domain data. Here, $x$ refers to an image, while $y=(b, c)$  denotes an annotation containing the bounding box $b$ and category $c$. UDA methods require labeled source data $D_s$ and unlabeled target data $D_t$ for adaptation. In contrast, SFOD settings utilize only a source-trained model $G$ (trained on labeled source data $D_s$) and the unlabeled target data $D_t$ for adaptation.

\subsection{Mean Teacher Self-training} 
\label{sec:mean_teacher}
The Mean Teacher (MT) framework \cite{Mt} consists of a teacher model $G_{\bar{\theta}}^{tea}$ and a student model $G_{\theta}^{stu}$ with identical architectures and initialization weights. The student model is optimized with pseudo labels provided by the teacher model, while the teacher model is progressively updated through EMA from the student model. In addition, the student and teacher models receive strong and weak augmentations. Enhanced consistency in their predictions improves the detection of target data \cite{Umt, Ut}.

Specifically, the student model is optimized using detection loss $\mathcal {L}_{det}$ computed from the student predictions ($\hat {b}$, $\hat {c}$) with corresponding pseudo labels ($b_t$, $c_t$) obtained from the teacher model after Hungarian matching, as detailed in \cite{Defdetr}:
\begin{equation}
   \mathcal {L}_{det} = \mathcal {L}_{cls}(\hat {c}, c_t) + \mathcal {L}_{L1}(\hat {b}, b_t) + \mathcal {L}_{giou}(\hat {b}, b_t)
  \label{eq:loss_t}
\end{equation}
where $\mathcal {L}_{cls}$ denotes the classification loss, $\mathcal {L}_{L1}$ and $\mathcal {L}_{giou}$ represent the regression losses.

The teacher model is progressively updated via EMA from the student model to enhance detection capability:
\begin{equation}
  \bar{\theta}_t \leftarrow \alpha \bar{\theta}_{t-1} + (1 - \alpha ) \theta_t 
  \label{eq:ema}
\end{equation}
where $\bar{\theta}_t$ and $\theta_t$ represent the parameters of the teacher and student models at the $t^{th}$ iteration, respectively, and $\alpha$ denotes a hyper-parameter update weight.

\subsection{Masked Image Modeling}
Because no ground truth is available in the target domain, utilizing noisy pseudo labels may result in model memorization. Encouraging the model to learn the context relations within the target domain images may mitigate the risk of memorization. Therefore, we adapted the Masked Image Consistency (MIC) \cite{Mic} module to facilitate context relations learning. MIC employs masked images to improve context learning for the target domain.
The masked target image $x^{M}$ is acquired through element-wise multiplication of the mask and the image:
\begin{equation}
   x^{M} = \mathcal {M} \odot x^{S} 
  \label{eq:mic}
\end{equation}
where $\mathcal {M}$ represents a patch mask generated from a uniform distribution, and $x^{S}$ denotes the strongly augmented target image from the student branch.

During the adaptation process, the masked image $x^{M}$ operates in the same way as the strongly augmented image $x^{S}$ within the student branch, as illustrated in \cref{fig:proposed}. Let $\mathcal {L}_{mask}$ represent the loss from $x^{M}$ and $\mathcal {L}_{strong}$ denote the loss from $x^{S}$. The total loss is defined as: 
\begin{equation}
  \mathcal {L}_{teach} = \mathcal {L}_{strong} + \mathcal {L}_{mask} 
  \label{eq:loss_teach}
\end{equation}
% where $\lambda _{mask}$ is set to 1 for simplicity.
\section{Proposed Method}
\subsection{Overview}
% Intuitively, a fundamental requirement for overcoming degradation in MT training is the co-evolution of both student and teacher modules, where the teacher can appropriately aggregate valuable knowledge from the student, and the student can actively overcome local optimums caused by incorrect pseudo labels. Based on observation, The student undergoes dynamic retraining " a local optimum during training, while the teacher model is dynamically updated based on the evolved retrained student.

Our proposed Dynamic Retraining-Updating (DRU) method aligns with the structure of the MT framework as described in \cref{sec:mean_teacher}. The overview of DRU is presented in \cref{fig:proposed}. DRU introduces a {\bf Dynamic Retraining-Updating mechanism} to actively manage student training and teacher updating based on the student's evolution. Specifically, we evaluate the student's progress by estimating its prediction uncertainty over historical data $D_{his}$. Depending on the received feedback, if the student has evolved, the teacher experiences {\bf Teacher Dynamic Updating}. In contrast, if the student has been trapped, it undergoes {\bf Student Dynamic Retraining}. Additionally, the student model receives further supervision from {\bf Historical Student Loss}, leveraging knowledge from the historical student model.
The details of the training process can be seen in \cref{alg:dru}.
\begin{algorithm}
\renewcommand{\algorithmicrequire}{\textbf{Input:}}
\renewcommand{\algorithmicensure}{\textbf{Output:}}
\caption{Dynamic Retraining-Updating training process}\label{alg:dru}
\begin{algorithmic}
\Require Teacher model $G_{\bar{\theta}}^{tea}$, student model $G_{\theta}^{stu}$, unlabeled target data $D_t$, uncertainty feedback $U$, and meta-iteration $M$.
\Ensure Optimized teacher model $G_{\bar{\theta}}^{tea}$
\State Empty buffer $D_{his}$; Copy $G_{\theta}^{stu}$ as $G_{\theta_{init}}^{stu}$; Index $i = 0$
\For{image batch $x_b$ in $D_t$}
\State Update $G_{\theta}^{stu}$ by \cref{eq:total_loss} with $x_b$, $G_{\bar{\theta}}^{tea}$ and $G_{\theta_{init}}^{stu}$ \Comment{with Historical Student Loss}
\State Append $x_b$ to $D_{his}$; $i++$
\If{$i<M$}
\If{$U[G_{\theta}^{stu}(D_{his})] < U[G_{\theta_{init}}^{stu}(D_{his})]$} \Comment{Student evolved}
\State Update $G_{\bar{\theta}}^{tea}$ by \cref{eq:dynamic_ema} with $G_{\bar{\theta}}^{tea}$ and $G_{\theta}^{stu}$ \Comment{Teacher Dynamic Updating}
\State Reset $D_{his}$; Copy $G_{\theta}^{stu}$ as $G_{\theta_{init}}^{stu}$; $i = 0$
\EndIf
\Else \Comment{Student trapped in a local optimum}
\State Reinitialize $G_{\theta}^{stu}$ with $G_{\theta}^{stu}$ and $G_{\theta_{init}}^{stu}$ \Comment{Student Dynamic Retraining}
\State Reset $D_{his}$; Copy $G_{\theta}^{stu}$ as $G_{\theta_{init}}^{stu}$; $i = 0$
\EndIf
\EndFor
\State \textbf{return} $G_{\bar{\theta}}^{tea}$
\end{algorithmic}
\end{algorithm}

\subsection{Dynamic Retraining-Updating Mechanism}
Intuitively, the co-evolution of student and teacher models is crucial for overcoming degradation, where the teacher appropriately integrates knowledge from the evolved student. Therefore, the evolution of the student plays a critical role. Inspired by research on semantic segmentation \cite{Tow}, which suggests that the performance of the student model on historical samples can reflect its evolution, we then aim to assess the student's progress by evaluating its performance over historical data $D_{his}$. In the scenario where target labels are unavailable for evaluation, we can assess the student's performance based on its prediction uncertainty \cite{Csfda, Tow}. Lower uncertainty in the student's output signifies higher adaptability to the target domain. To estimate the student's prediction uncertainty, we leverage the inherent structure of the decoder, where the variation of outputs across all decoder layers could give us a close estimate \cite{Caldetr}. Specifically, let consider $O_D \in \mathbb{R}^{L \times B \times Q \times C}$ as the outputs of all decoder layers, where $L$, $B$, $Q$, and $C$ denote the number of decoder layers, batch size, number of queries, and number of classes, respectively. By computing the variance along the $L$ dimension in $O_D$, we yield $u \in \mathbb{R}^{B \times Q \times C}$, which represents the student's prediction uncertainty for an image batch $x_b \in D_{his}$. Based on the feedback $U=\{u\}$, we actively update the teacher and retrain the student, as described in \cref{alg:dru}.

\subsubsection{Student Dynamic Retraining}
Due to the lack of ground truths, the student model is only updated with noisy pseudo labels, which always contain bias caused by domain shifts. Without corrective measures, the student replicates this bias during training, making it vulnerable to being stuck in a sub-optimal state. Consequently, the teacher model could also be affected as it is continuously updated from the student model, generating even worse pseudo labels. To overcome this, we introduce dynamically selective retraining of the student model.

Specifically, during a training period $M$, if the student model cannot escape a downward trend, we selectively retrain parts of the student model. Despite being trapped in a local optimum due to noisy pseudo labels, the current student model has acquired valuable feature knowledge from the target domain during learning, thereby benefiting the detector. Therefore, while continuously updating parameters of the backbone and encoder modules throughout training progress, we re-initialize the decoder of the current student model $G_{\theta_{t}}^{stu}$ with the weights of the decoder from the historical student model $G_{\theta_{t-M}}^{stu}$. In this way, the retrained decoder can leave the local optimum, guided by better-trained backbone and encoder components. We verify this assumption in ablation studies.

\subsubsection{Teacher Dynamic Updating}
The teacher model can be viewed as an ensemble of multiple temporal student models. Uncontrolled updates without regard to student conditions can lead to accumulating errors in the teacher model, making it susceptible to detection deterioration. As the student model acquires new knowledge and becomes more reliable to the target domain, updating the teacher model facilitates the integration of valuable insights from the student model, thereby enhancing its detection capability. Let $\tilde{m}$ denote the dynamic update interval. The EMA described in \cref{eq:ema} is reformulated as follows:
\begin{equation}
  \bar{\theta} _t \leftarrow \alpha \bar{\theta} _{t-\tilde{m}} + (1 - \alpha ) \theta _t 
  \label{eq:dynamic_ema}
\end{equation}

% \begin{equation}
%   G_{\theta_{\text{s}}^{\text{t}}}^{\text{stu}} = {\Large\bf Reinitialize}(G_{\theta_{\text{s}}^{\text{t}}}^{\text{stu}}, G_{\theta_{\text{s}}^{\text{t-m}}}^{\text{stu}}, keep\_params= \text{[backbone, encoder]})
%   \label{eq:re_init}
% \end{equation}

\subsection{Historical Student Loss}
In DRU, the historical student acts as an additional teacher, setting a performance lower bound for updating the current student model. Suppose the student model faces a severe performance decrease due to a deteriorating teacher model. In that case, the historical student ensures that the student model maintains a forward-reaching trajectory, thereby preventing rapid performance deterioration.

Due to inaccuracies in predicted locations and to prevent inconsistencies with pseudo labels bounding box from the teacher model, the Historical Student Loss $\mathcal {L}_{his}$ includes only the classification task: 
\begin{equation}
 \mathcal {L}_{his} = \mathcal {L}_{cls}(\hat {c}, {c}_{his} )
  \label{eq:loss_his}
\end{equation}

To mimic the historical performance and reduce the influence of incorrect pseudo labels, $\mathcal {L}_{cls}$ is adjusted as follows, inspired by Quality Focal Loss \cite{Gfl}:
\begin{equation}
 \mathcal {L}_{cls}(\hat {c}, {c}_{his}) = \sum _{i=1}^{N_{pos}} \left |{s}_{i}-\hat {c_i}\right |^{\gamma } B C E\left (\hat {c_i}, {s}_{i}\right )+\sum _{j=1}^{N_{neg}} \hat {c}_{j}^{\gamma } B C E\left (\hat {c_j}, 0\right ) 
  \label{eq:loss_cls}
\end{equation}
where $\hat {c_i}$, ${s}_{i}$ represent the predicted probability of current student $G_{\theta_{t}}^{stu}$ and the corresponding confidence score of class ${c}_{his}$ from historical student $G_{\theta_{his}}^{stu}$ of the $i^{th}$ example, respectively (More details are included in supplementary material).

% of $i^{th}$ examples
% $N_{pos}$ and $N_{neg}$ denote the number of positive and negative examples, $BCE$ stands for binary cross-entropy, $\gamma$ is focusing parameter (default 2).

The $\mathcal {L}_{his}$ operates in the same way as $\mathcal {L}_{teach}$ in \cref{eq:loss_teach}. For simplicity, the overall objective loss for the student model is:
\begin{equation}
  \mathcal {L}_{total} = \mathcal {L}_{teach} + \mathcal {L}_{his}
  \label{eq:total_loss}
\end{equation}
% where $\lambda _{his}$ is simply set to 1.
\section{Experiments}
\subsection{Experimental Setup}
\subsubsection{Datasets} Following the setting of existing SFOD \cite{Sed, A2sfod, Pets} and UDA \cite{Sfa, Mttrans, O2net} methods, we evaluate the performance of DRU in three popular domain adaptation benchmarks from four public datasets: Cityscapes \cite{Cityscapes}, Foggy Cityscapes \cite{Foggy}, Sim10k \cite{Sim10k}, and BDD100k \cite{Bdd100k}.
\begin{itemize}
\item {\bf Normal to Foggy Adaptation} Cityscapes dataset is gathered from urban environments. It includes 3,475 annotated images, with 2,975 allocated for training and the remaining 500 reserved for evaluation. Foggy Cityscapes is generated from Cityscapes using a fog synthesis algorithm. We employ Cityscapes as the source domain and Foggy Cityscapes with the highest fog density (0.02) as the target domain.
\item {\bf Cross Scene Adaptation} BDD100k is a large-scale driving dataset. We employ the BDD100k daytime subset as the target domain, consisting of 36,728 training images and 5,258 validation images with ground truth. The Cityscapes remains as the source domain.
\item {\bf Synthetic to Real Adaptation} Sim10k comprises 10,000 synthetic images generated from the GTA game engine. We utilize Sim10k as the source data and ``car'' instances from Cityscapes as the target data.
% \item {\bf Cross Camera Adaptation.} KITTI comprises 7,481 training images captured from various real-world street scenes. We employ KITTI as the source domain and "car" instances from Cityscapes as the target domain.
\end{itemize}

\subsubsection{Implementation Details} Our method is built upon Deformable DETR \cite{Defdetr}. We set the smoothing parameter $\alpha = 0.999$ for the EMA ($\alpha = 0.9996$ for Cityscapes $\to$ BDD100k). The threshold is defined as $0.3$ for filtering pseudo labels. A meta-iteration for Dynamic Retraining-Updating is set to $M=5$. The network is optimized with the Adam optimizer with an initial learning rate of $2 \times 10^{-4}$. The batch size is fixed at $8$ for all domain adaptation scenarios. The data augmentation techniques comprise random horizontal flipping for weak augmentation and random color jittering, grayscaling, and Gaussian blurring for strong augmentation \cite{Mrt}. For the MIC module, we use a patch size $b=64$ and a mask ratio $r=0.5$ as the objects of interest. We utilize Mean Average Precision (mAP) with a threshold of $0.5$ as the evaluation metric. All experiments are conducted on an NVIDIA Quadro RTX 8000 GPU (48GB).

\begin{table}[tb]
  \caption{Results of different UDA and SFOD methods for Normal to Foggy Adaptation (Cityscapes $\to$ Foggy Cityscapes). ``Source Only'' refers to the source-trained model. FRCNN denotes Faster R-CNN, and DefDETR represents Deformable DETR.}
  \label{tab:weather}
  \centering
  \begin{tabular}{@{}c|c|c|cccccccc|c@{}}
    \toprule
    &Method & Detector & person & rider & car & truck & bus & train & motor & bicycle & mAP\\
    \midrule
    \midrule
    %&Source Only & FRCNN & 26.9 & 38.2 & 35.6 & 18.3 & 32.4 & 9.6 & 25.8 & 28.6 & 26.9 \\
    &Source Only & DefDETR & 40.0 & 41.2 & 47.0 & 13.0 & 29.1 & 6.5 & 21.5 & 38.0 & 29.5 \\
    \midrule
    \multirow{8}{*}{\rotatebox[origin=c]{90}{UDA}}
    &SW-Faster \cite{Swfaster} & FRCNN &  32.3 & 42.2 & 47.3 & 23.7 & 41.3 & 27.8 & 28.3 & 35.4 & 34.8 \\
    &CR-DA-DET \cite{Crdadet} & FRCNN & 32.9 & 43.8 & 49.2 & 27.2 & 45.1 & 36.4 & 30.3 & 34.6 & 37.4 \\
    &TIA \cite{Tia} & FRCNN & 34.8 & 46.3 & 49.7 & 31.1 & {\bf 52.1} & {\bf 48.6} & {\bf 37.7} & 38.1 & 42.3 \\
    &PT \cite{Pt} & FRCNN & 40.2 & 48.8 & 59.7 & 30.7 & 51.8 & 30.6 & 35.4 & 44.5 & 42.7 \\
    &TDD \cite{Tdd} & FRCNN & 39.6 & 47.5 & 55.7 & {\bf 33.8} & 47.6 & 42.1 & 37.0 & 41.4 & 43.1 \\
    &SFA \cite{Sfa} & DefDETR & 46.5 & 48.6 & 62.6 & 25.1 & 46.2 & 29.4 & 28.3 & 44.0 & 41.3 \\
    &MTTrans \cite{Mttrans} & DefDETR & 47.7 & 49.9 & {\bf 65.2} & 25.8 & 45.9 & 33.8 & 32.6 & 46.5 & 43.4 \\
    &DA-DETR \cite{Dadetr} & DefDETR & {\bf 49.9} & 50.0 & 63.1 & 24.0 & 45.8 & 37.5 & 31.6 & 46.3 & 43.5 \\
    \midrule
    \multirow{6}{*}{\rotatebox[origin=c]{90}{SFOD}}&SED(Mosaic) \cite{Sed} & FRCNN & 33.2 & 40.7 & 44.5 & 25.5 & 39.0 & 22.2 & 28.4 & 34.1 & 33.5\\
    %&HCL \cite{Hcl} & FRCNN & 26.9 & 46.0 & 41.3 & {\bf 33.0} & 25.0 & 28.1 & 35.9 & 40.7 & 34.6 \\
    &A${^2}$SFOD \cite{A2sfod} & FRCNN & 32.3 & 44.1 & 44.6 & 28.1 & 34.3 & 29.0 & 31.8 & 38.9 & 35.4 \\
    &LODS \cite{Lods} & FRCNN & 34.0 & 45.7 & 48.8 & 27.3 & 39.7 & 19.6 & 33.2 & 37.8 & 35.8 \\
    &PETS \cite{Pets} & FRCNN & 42.0 & 48.7 & 56.3 & 19.3 & 39.3 & 5.5 & 34.2 & 41.6 & 35.9 \\
    &IRG \cite{Irg} & FRCNN & 37.4 & 45.2 & 51.9 & 24.4 & 39.6 & 25.2 & 31.5 & 41.6 & 37.1 \\
    &Ours & DefDETR & 48.3 & {\bf 51.5} & 62.5 & 26.2 & 43.2 & 34.1 & 34.2 & {\bf 48.6} & {\bf 43.6} \\
  \bottomrule
  \end{tabular}
\end{table}

\subsection{Comparison with SOTA Methods}
This section evaluates the proposed DRU model across three challenging domain adaptation scenarios and compares the results with other recent SOTA SFOD and UDA methods.
\subsubsection{Normal to Foggy Adaptation} Weather conditions frequently fluctuate, presenting significant challenges. Object detectors must maintain reliability under all circumstances. Therefore, we evaluate the detectors' robustness to weather changes by transitioning from Cityscapes to Foggy Cityscapes. As indicated in \cref{tab:weather}, DRU surpasses advanced SOTA SFOD methods by a considerable margin (43.6\% compared to the closest SOTA 37.1\%).

\begin{table}[tb]
  \caption{Results of Cross Scene Adaptation (Cityscapes $\to$  BDD100k)}
  \label{tab:scene}
  \centering
  \begin{tabular}{@{}c|c|c|c c c c c c c|c@{}}
    \toprule
    &Method & Detector & truck & car & rider & person & motor & bicycle & bus & mAP \\
    \midrule
    \midrule
    %&Source Only & FRCNN & 17.9 & 44.1 & 25.4 & 28.8 & 13.9 & 22.4 & 16.1 & 24.1 \\
    &Source Only & DefDETR & 18.9 & 58.2 & 28.3 & 42.0 & 15.7 & 18.8 & 21.7 & 29.1 \\
    \midrule
    \multirow{7}{*}{\rotatebox[origin=c]{90}{UDA}}
    &DA-Faster \cite{Dafaster} & FRCNN & 14.3 & 44.6 & 26.5 & 29.4 & 15.8 & 20.6 & 16.8 & 24.0 \\
    &SW-Faster \cite{Swfaster}& FRCNN & 15.2 & 45.7 & 29.5 & 30.2 & 17.1 & 21.2 & 18.4 & 25.3 \\
    &CR-DA-DET \cite{Crdadet} & FRCNN & 19.5 & 46.3 & 31.3 & 31.4 & 17.3 & 23.8 & 18.9 & 26.9 \\
    &AQT \cite{Aqt} & DefDETR & 17.3 & 58.4 & 33.0 & 38.2 & 16.9 & 23.5 & 18.4 & 29.4 \\
    &O$^2$net \cite{O2net} & DefDETR & 20.4 & 58.6 & 31.2 & 40.4 & 14.9 & 22.7 & 25.0 & 30.5 \\
    &MTTrans \cite{Mttrans} & DefDETR & 25.1 & 61.5 & 30.1 & 44.1 & 17.7 & 23.0 & 26.9 & 32.6 \\
    &MRT \cite{Mrt} & DefDETR & 24.7 & {\bf 63.7} & 30.9 & {\bf 48.4} & 20.2 & 22.6 & 25.5 & 33.7 \\
    \midrule
    \multirow{4}{*}{\rotatebox[origin=c]{90}{SFOD}}
    % &SED \cite{Sed} & FRCNN & 20.4 & 48.8 & 32.4 & 31.0 & 15.0 & 24.3 & 21.3 & 27.6 \\
    &SED(Mosaic) \cite{Sed} & FRCNN & 20.6 & 50.4 & 32.6 & 32.4 & 18.9 & 25.0 & 23.4 & 29.0 \\
    &PETS \cite{Pets} & FRCNN & 19.3 & 62.4 & 34.5 & 42.6 & 17.0 & 26.3 & 16.9 & 31.3 \\
    &A${^2}$SFOD \cite{A2sfod} & FRCNN & 26.6 & 50.2 & 36.3 & 33.2 & 22.5 & 28.2 & 24.4 & 31.6 \\
    &Ours & DefDETR & {\bf 27.1} & 62.7 & {\bf 36.9} & 45.8 & {\bf 22.7} & {\bf 32.5} & {\bf 28.1} & {\bf 36.6} \\
  \bottomrule
  \end{tabular}
\end{table}

\subsubsection{Cross Scene Adaptation} Scene configurations dynamically change in real-world applications, especially in automated driving contexts. Hence, the model's performance in cross-scene adaptation is crucial. As depicted in \cref{tab:scene}, DRU achieves SOTA results compared to recent SFOD and UDA methods with substantial improvements, surpassing the previous SOTA method by 5.0\%. Furthermore, it enhances the results of five out of seven classes in the target domain.

\begin{table}[tb]
  \centering
  \begin{minipage}[t]{0.48\linewidth}
     \caption{Results of Synthetic to Real Adaptation (Sim10k $\to$  Cityscapes)}
     \label{tab:synthetic}
     \centering
     \begin{tabular}{@{}c|c|c|c@{}}
         \toprule
         &Method & Detector & AP of car \\
         \midrule
         \midrule
         %&Source Only & FRCNN & 39.4 \\
         &Source Only & DefDETR & 48.9 \\
         \midrule
        \multirow{7}{*}{\rotatebox[origin=c]{90}{UDA}}
         &TDD \cite{Tdd} & FRCNN & 53.4 \\
         &PT \cite{Pt} & FRCNN & 55.1 \\
         &SFA \cite{Sfa} & DefDETR & 52.6 \\
         % &AQT \cite{Aqt} & DefDETR & 53.4 \\
         &O$^2$net \cite{O2net} & DefDETR & 54.1 \\
         &DA-DETR \cite{Dadetr} & DefDETR & 54.7 \\
         &MTTrans \cite{Mttrans} & DefDETR & 57.9 \\
         &MTM \cite{Mtm} & DefDETR & 58.1 \\
         \midrule
         \multirow{5}{*}{\rotatebox[origin=c]{90}{SFOD}}
         &SED(Mosaic) \cite{Sed} & FRCNN & 43.1 \\ 
         &IRG \cite{Irg} & FRCNN & 43.2 \\ 
         &A${^2}$SFOD \cite{A2sfod} & FRCNN & 44.0 \\
         &PETS \cite{Pets} & FRCNN & 57.8 \\
         &Ours & DefDETR & {\bf 58.7} \\
        \bottomrule
     \end{tabular}
  \end{minipage}%
  \hfill
  \hfill
  \begin{minipage}[t]{0.44\linewidth}
     \caption{Ablation studies of adding modules to MT framework on Cityscapes $\to$ Foggy Cityscapes. ``Src'' denotes the Source Only trained model. ``MT'' represents the Mean Teacher baseline. ``MIC'', ``$\mathcal {L}_{his}$'', and ``DRU'' denote the Masked Image Modeling, Historical Student Loss, and Dynamic Retraining-Updating, respectively.}
     \label{tab:ablation_modules}
     \centering
      \begin{tabular}{@{} c c  c c c |c  c @{}}
        \toprule
        Src & MT & MIC & $\mathcal {L}_{his}$ & DRU & mAP & gain \\
        \midrule
        $\checkmark$ &  &  &  &    & 29.5 &   \\
        $\checkmark$ & $\checkmark$ &  &  &    & 37.4 & $+$7.9  \\
        $\checkmark$ & $\checkmark$ & $\checkmark$ &  &    & 39.8 & $+$10.3  \\
        $\checkmark$ & $\checkmark$ & $\checkmark$ & $\checkmark$ &    & 41.3 & $+$11.8  \\
        $\checkmark$ & $\checkmark$ & $\checkmark$ &  & $\checkmark$ &   40.9 & $+$11.4 \\
        $\checkmark$ & $\checkmark$ & $\checkmark$ & $\checkmark$ & $\checkmark$ &   {\bf 43.6} & {\bf $+$14.1} \\
      \bottomrule
    \end{tabular}
  \end{minipage}%
\end{table}

\subsubsection{Synthetic to Real Adaptation} 
Synthetic images and labels can be automatically generated, providing an alternative to address the challenges of data collection and labeling. Therefore, enabling the model to transfer from synthetic to real images is advantageous. As shown in \cref{tab:synthetic}, we evaluate DRU in the synthetic to real adaptation scenario, achieving a SOTA performance of 58.7\%. DRU surpasses the latest SOTA SFOD method by 0.9\% and is even comparable with recent SOTA UDA methods.
% \subsubsection{Cross Camera Adaptation.} Domain shifts are inevitable in cross-camera images due to differences in camera configurations. The detector needs to achieve desirable results in handling shifts in cross-camera images. The results presented in Table 4 indicate that our method achieves state-of-the-art performance on this benchmark.

\subsection{Ablation Studies} 
\subsubsection{Modules Analysis}
To evaluate the effectiveness of each component in our method, we conduct ablation studies by progressively incorporating individual components into DRU. \Cref{tab:ablation_modules} presents some observations: \emph{(1)} Integrating the baseline MT framework directly into a source-trained model leads to remarkable progress in the target domain (+7.9\%). This emphasizes the significance of the MT framework in self-training adaptation, yet there is still room for further enhancement. \emph{(2)} The incorporation of MIC into MT results in a performance increase of 10.3\%, attributed to its effective learning of context relations in the target domain. \emph{(3)} Using additional Historical Student Loss reduces the influence of incorrect pseudo labels, resulting in an 11.8\% gain. \emph{(4)} Finally, incorporating a Dynamic Retraining-Updating mechanism effectively manages teacher updating and student training, resulting in a 14.1\% improvement. 

\subsubsection{Dynamic Retraining-Updating Analysis}
We analyze the usefulness of selective retraining. When the student model becomes trapped in a local optimum, we selectively re-initialize specific components while preserving others as enhanced parts. \Cref{tab:dru-a} shows that maintaining updates to the backbone and encoder while re-initializing the decoder produces the optimal result. This result suggests that the model has acquired meaningful feature representations in the backbone and encoder, whereas the decoder's performance deteriorates due to incorrect pseudo labels. We also evaluate the effectiveness of fixed and dynamic update intervals for the teacher model in \cref{tab:dru-b}. With appropriate dynamic updating, the teacher model can effectively aggregate the valuable knowledge from the evolved student model, achieving the best result. Additionally, it addresses the inconvenience and complication of manually searching for optimal hyper-parameters. We examine various values of meta-iteration $M$, which is shown in \cref{tab:dru-c}, and find that the model yields the best performance with $M=5$.  When $M$ is too small or too large, the model fails to update efficiently.

% \begin{table}[tb]
%   \centering
%   % \caption{Car}
%   % \label{tab:car}
%   \begin{minipage}[t]{0.5\linewidth}
%      \caption{Ablation studies with selective retraining on Cityscapes $\to$ Foggy Cityscapes. "Bac+Enc" denotes retrained Backbone and Encoder, "Dec" represents retrained Decoder, and "No" indicates keep training all modules.}
%      \label{tab:retrained}
%      \centering
%      \begin{tabular}{@{}c | c c c@{}}
%         \toprule
%          Retrained  & No & Bac+Enc & Dec \\
%         \midrule
%          mAP & 41.2 & 39.9 & {\bf 43.6} \\
%      \bottomrule
%      \end{tabular}
%   \end{minipage}%
%   \hfill
%   \hfill
%   \begin{minipage}[t]{0.4\linewidth}
%      \caption{Ablation studies with varying $m$ on Cityscapes $\to$ Foggy Cityscapes.}
%      \label{tab:interval}
%      \centering
%      \begin{tabular}{@{}c | c c c c@{}}
%         \toprule
%          $m$ & 2 & 5 & 10 & 20\\
%         \midrule
%          mAP & 41.9 & {\bf 43.6} & 42.6 & 41.5\\
%      \bottomrule
%      \end{tabular}
%   \end{minipage}%
% \end{table}

\begin{table}[tb]
  \caption{Ablation studies of Dynamic Retraining-Updating on Cityscapes $\to$ Foggy Cityscapes. (a) Selective retraining, ``Bac+Enc'' denotes retrained Backbone and Encoder, ``Dec'' means retrained Decoder, and ``No'' indicates keeping training all modules. (b) Dynamic updating, ``$\tilde{m}$'' means dynamic update interval. (c) Meta-iteration $M$.}
  \label{tab:dru}
  \centering
  \begin{subtable}{0.36\linewidth}
     \caption{Selective retraining}
     \label{tab:dru-a}
     \centering
     \begin{tabular}{@{}c | c c c@{}}
        \toprule
         Retrained  & No & Bac+Enc & Dec \\
        \midrule
         mAP & 41.2 & 39.9 & {\bf 43.6} \\
     \bottomrule
     \end{tabular}
  \end{subtable}%
  \begin{subtable}{0.34\linewidth}
     \caption{Dynamic updating}
     \label{tab:dru-b}
     \centering
     \begin{tabular}{@{}c | c c c c@{}}
        \toprule
         $i$ & 1 & 5 & 10 & $\tilde{m}$\\
        \midrule
         mAP & 41.8 & 43.4 & 41.0 & {\bf 43.6}\\
     \bottomrule
     \end{tabular}
  \end{subtable}%
  \begin{subtable}{0.3\linewidth}
     \caption{Meta-iteration $M$}
     \label{tab:dru-c}
     \centering
     \begin{tabular}{@{}c | c c c c@{}}
        \toprule
         $M$ & 2 & 5 & 10 & 20\\
        \midrule
         mAP & 41.9 & {\bf 43.6} & 42.6 & 41.5\\
     \bottomrule
     \end{tabular}
  \end{subtable}

\end{table}
\subsection{Result Analysis}
\subsubsection{Training Stability}
\begin{figure}[tb]
  \centering
  \includegraphics[width=1.0\columnwidth]{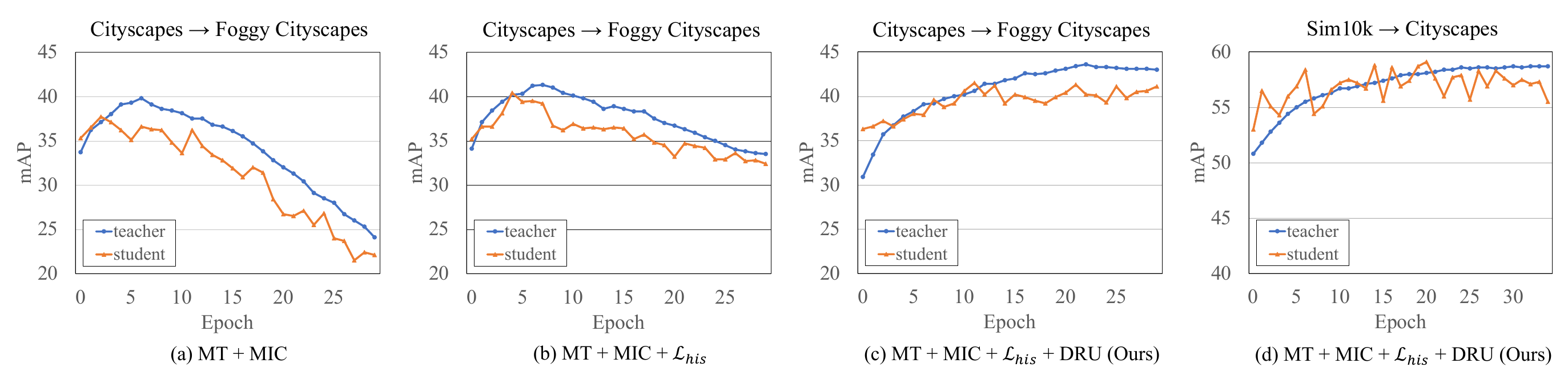}
  \caption{(a), (b), (c) The training curves for adding modules to MT on Cityscapes $\to$ Foggy Cityscapes. (d) The training curves of our method on Sim10k $\to$  Cityscapes.
  }
  \label{fig:module_curve}
\end{figure}

To validate the effectiveness of our DRU in training stability, we present the mAP curves of the student and teacher models on the validation dataset throughout training. As shown in \cref{fig:module_curve} (a), the mAP curves of both student and teacher models in conventional MT initially steadily increase but rapidly degrade afterward, indicating the poor stability of the foundational self-training MT approach. In \cref{fig:module_curve} (b), the mAP curves, supplemented with Historical Student Loss, exhibit a steady initial rise followed by a slight decrease, highlighting the effectiveness of Historical Student Loss in mitigating severe performance degradation caused by incorrect pseudo labels. Finally, in \cref{fig:module_curve} (c), incorporating the Dynamic Retraining-Updating mechanism results in accelerated improvement and enhanced stability, demonstrating its capability to promote the co-evolution of student and teacher models through the training process. Furthermore, as illustrated in \cref{fig:module_curve} (c), (d), similar training curves across different domain adaptation scenarios confirm the effectiveness of our method in stabilizing self-training.

\begin{figure}[tb]
  \centering
  \includegraphics[width=1.0\columnwidth]{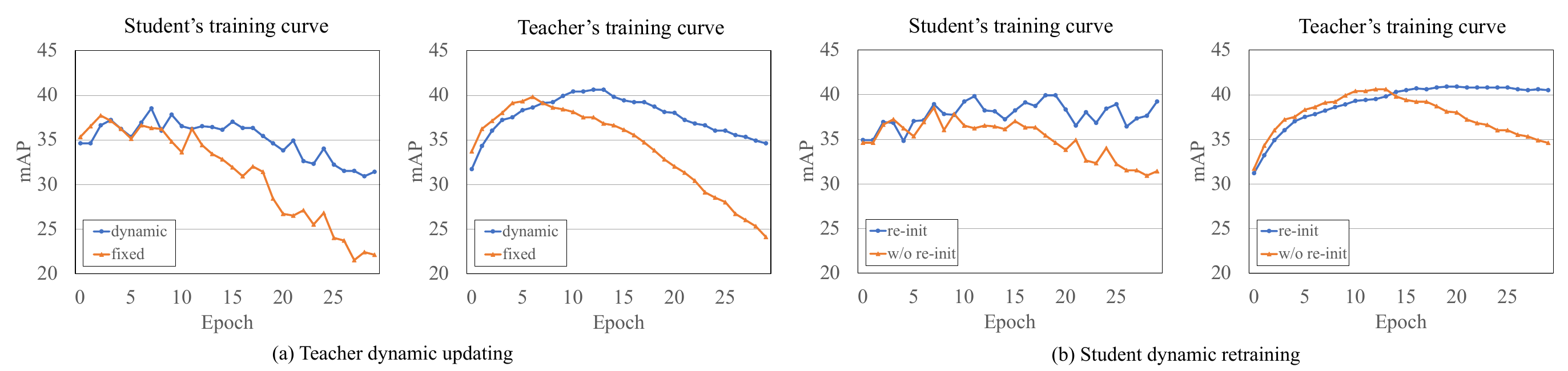}
  \caption{The training curves of student and teacher models with Dynamic Retraining-Updating mechanism on Cityscapes $\to$ Foggy Cityscapes 
  }
  \label{fig:retrain_curve}
\end{figure}

We further analyze the stability of dynamically retraining the student model and updating the teacher model without using Historical Student Loss. As depicted in \cref{fig:retrain_curve} (a), appropriate dynamic updates of the teacher model contribute to training stability and later enhance performance. However, the student cannot escape the downward trend, resulting in a subsequent decrease in performance. In \cref{fig:retrain_curve} (b), with additional dynamic parameter re-initialization, the student model can converge to a better local optimum, leading to gradual performance improvements, particularly in the later stages of the training process.  

\subsubsection{Visualization}
\begin{figure}[tb]
  \centering
  \includegraphics[width=1.0\columnwidth]{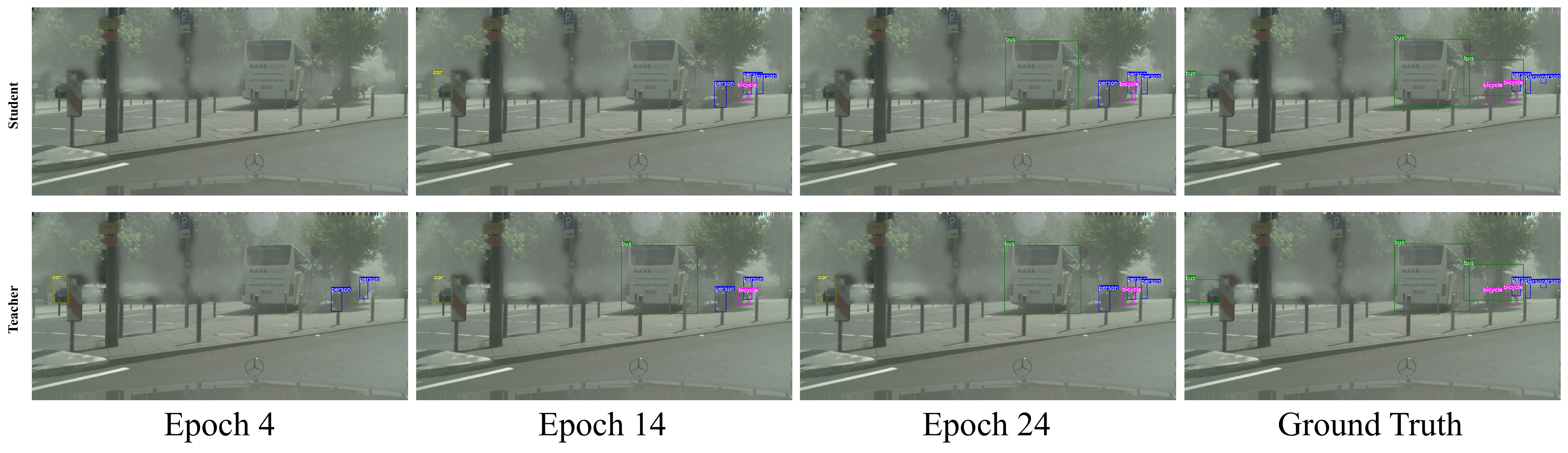}
  \caption{Detection results of our student and teacher models at different epochs on Cityscapes $\to$ Foggy Cityscapes. Our DRU shows the gradual improvement of both the student and teacher models through the adaptation process.
  }
  \label{fig:vis_epoch}
\end{figure}

\begin{figure}[tb]
  \centering
  \includegraphics[width=1.0\columnwidth]{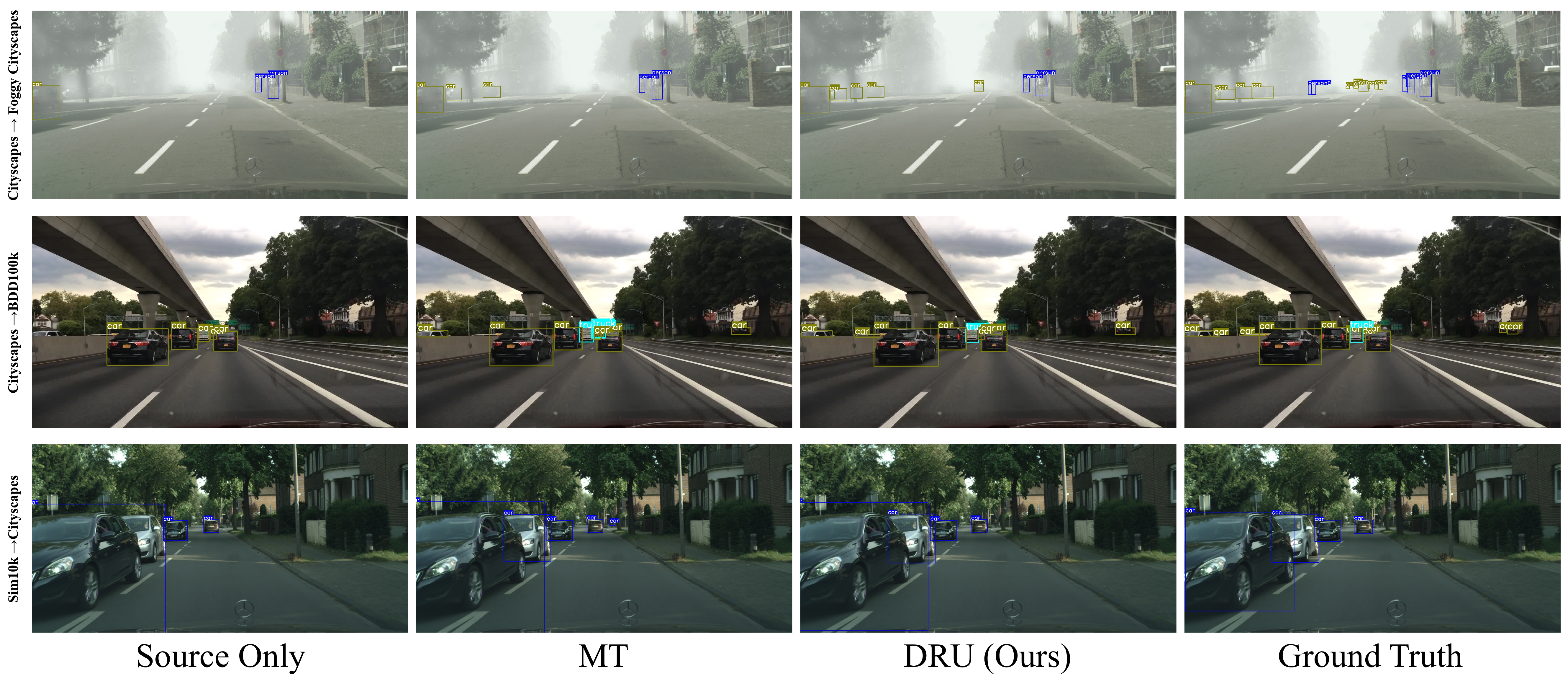}
  \caption{Detection results of our DRU on different domain adaptation scenarios
  }
  \label{fig:vis_various}
\end{figure}

In \cref{fig:vis_epoch}, the predictions of the student and teacher models are visually depicted at various training epochs in the Cityscapes $\to$ Foggy Cityscapes task. Initially, there were missing predictions for objects. However, as training progressed, more objects were detected, and the predictions gradually became more accurate. This observation confirms the effectiveness of our method in preventing the deterioration phenomenon in self-training and achieving co-evolution in both student and teacher models through training progress. We further illustrate the predictions of our method on more challenging adaptation tasks, as depicted in \cref{fig:vis_various}. The positive results suggest that our approach performs properly in many self-training adaptation scenarios, even with limited source-trained models.

\section{Conclusion}
This paper focuses on addressing the challenges of domain adaptive object detection in scenarios where the source domain is unavailable. We investigate the causes of the deterioration issue in the self-training Mean Teacher framework and propose relevant enhancement approaches. Specifically, we introduce the Dynamic Retraining-Update mechanism to promote the co-evolution in student and teacher models. Also, we propose the Historical Student Loss to mitigate the impact of noisy pseudo labels. Across various SFOD benchmarks, our method significantly enhances the stability and adaptability of the self-training paradigm, achieving state-of-the-art performances that are even comparable to advanced UDA methods. We expect our research can provide new insights and improve the effectiveness of self-training methods in more intricate scenarios.

% \clearpage\mbox{}Page \thepage\ of the manuscript. This is the last page.
\par\vfill\par

% \clearpage  % TODO REVIEW/FINAL: This \clearpage needs to be removed from both review and camera-ready versions.
\section*{Acknowledgements}
This work was supported by the Institute of Information \& Communications Technology Planning \& Evaluation(IITP) grant funded by the Korea government(MSIT) (No. 2021-0-01364, An intelligent system for 24/7 real-time traffic surveillance on edge devices)

\bibliographystyle{splncs04}
\bibliography{egbib}
\end{document}